# k-Parameter Approach for False In-Season Anomaly Suppression in Daily Time Series Anomaly Detection

**Vincent Yuansang Zha[1], Vaishnavi Kommaraju[1], Okenna Obi-Njoku[1], Vijay Dakshinamoorthy[1], Anirudh Agnihotri[1], Nantes Kirsten[1]**

[1]Bell Canada

*Abstract*—Detecting anomalies in a daily time series with a weekly pattern is a common task with a wide range of applications. A typical way of performing the task is by using decomposition method. However, the method often generates false positive results where a data point falls within its weekly range but is just off from its weekday position. We refer to this type of anomalies as "in-season anomalies", and propose a k-parameter approach to address the issue. The approach provides configurable extra tolerance for in-season anomalies to suppress misleading alerts while preserving real positives. It yields favorable result.

## I. INTRODUCTION

Daily anomaly detection is a common task with broad applications, and the decomposition method is widely used to perform the task. However, it often generates false positives due to over-fitting of the weekly pattern.

The remainder of this paper is organized as follows: Section 2 discusses the limitations of the classical decomposition method, and introduces the concept of "in-season anomalies" which often cause false positives. Section 3 proposes a solution referred to by this paper as the "k-parameter approach". Section 4 outlines the positive results. Section 5 draws the conclusion and provides suggestions for further studies.

## II. PROBLEM STATEMENT

### A. Review of classical decomposition method

Detecting anomalies in a daily time series with a weekly periodicity is a common task with broad applications. For example, a usage or revenue stream in a line of business on a daily basis usually contains a weekly pattern. Detecting anomalies against its weekly pattern can assist stakeholders in receiving alerts and taking timely action.

A typical way of performing the task is by using decomposition method. The method treats a series as a combination of trend, seasonality (weekly pattern in the case of daily time series), and residual components, and utilizes the residual component as a new representation for detection [1, 2, 3, 4]. The procedure can be summarized by the following formulas.

$$\begin{aligned}
v_p &= trend \cdot season \\
resid &= \frac{v}{v_p} \\
thresh &= n \cdot \text{std}(resid) \\
class &= 0, if\ resid \in [\frac{1}{1+thresh}, 1 + thresh] \\
class &= 1, if\ otherwise
\end{aligned} \quad (1)$$

where

$v_p$ is the predicted value;
$trend$ is the trend component;
$season$ is the seasonality component;
$resid$ is the residual component;
$v$ is the actual value of the data point;
$thresh$ is the threshold of anomaly detection;
$n$ is the number of standard deviations. It serves as the anomaly criteria;
std is the standard deviation function;
$class$ is the classification result of the anomaly detection. $class = 0$ means non-anomaly, and $class = 1$ means anomaly.

The formulas are based on the premise of multiplicative growth of the time series. They are easily adaptable to additive assumption, which will not be covered by this paper.

Some improvements are made, such as enhancing decomposition based on loess [5], applying exponential smoothing method [6], employing regression method [7], boosting resilience for long time series decomposition [8], and using auto-encoder method [9].

To illustrate the method, we created a simulator to generate time series having a weekly pattern with certain randomness and anomalies. Fig 1 is a typical example of the time series generated by the simulator.

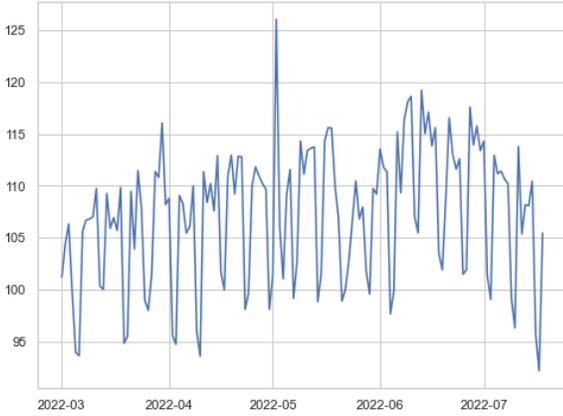

Fig 1. Original daily time series. It is synthesized by a simulation. It has a strong weekly pattern.

Fig 2 depicts the decomposition method against the time series. Fig 2 (a) is the original series. It is decomposed into three components: trend in Fig 2 (b), seasonality in Fig 2 (c), and residual in Fig 2 (d).

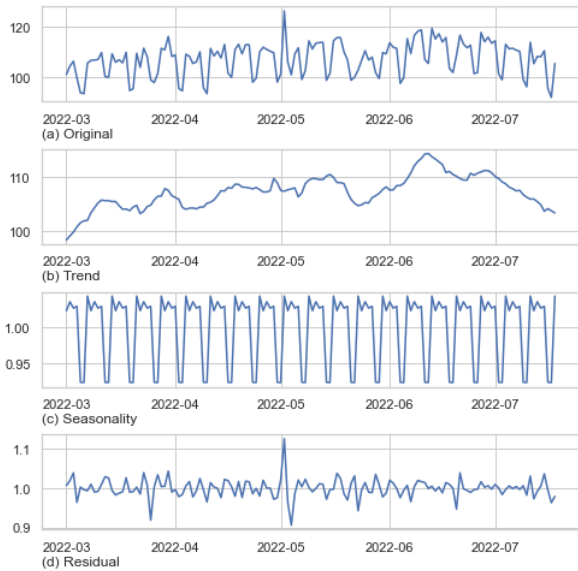

Fig 2. Time series decomposition. (a) is the original time series having an apparent weekly pattern. It is decomposed into three components: trend in (b), seasonality in (c), and residual in (d).

Based on the residual in Fig 2 (d), anomalies can be determined according to z-score, assuming the residuals largely follows a normal distribution. If the residual's z-score of any data point is outside of a specified limit, for example, if it is greater than 2, the model can conclude that the data point is an anomaly.

Fig 3 depicts the detection result of the daily series illustrated in Fig 1. The model detects five anomalies that are highlighted with red circles: A, B, C, D, and E. It is based on a threshold of $z = 2.00$, i.e. 2.00 standard deviation, or 2.00 STD. The threshold is selected such that the resulting anomalies are neither too few nor too many. The decision boundary is illustrated by the gray area.

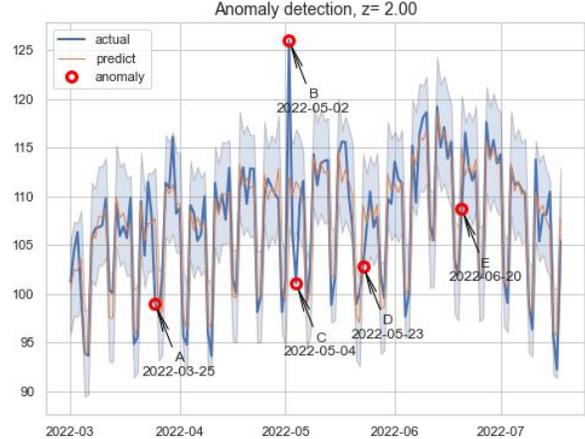

Fig 3. Outcome of anomaly detection using the classical decomposition method. The blue line represents the actual value. The orange line represents the predicted value. The decision boundary is shown by the gray area, which is set by a z=2.00 threshold. The model detects five anomalies that are highlighted with red circles: A, B, C, D, and E. In particular, A, C, D, and E are in-season anomalies, and B is a non-in-season anomaly.

### B. Introduction of the concept of the "in-season anomaly" as a problem statement

In Fig 3, however, while certain anomalies, such as B, are clear, others, such as D and E, do not appear obvious. Those non-obvious anomalies share the characteristics: the values of the data points are within the week's range, but they are "off" in terms of their particular weekdays.

Consider E as an example; it is apparently inside its weekly range. In other words, certain neighbouring days have values greater than E, and some others less than E. Therefore, E falls inside the range between the minimum and maximum values of its week, and hence does not stand out as an "anomaly" relative to its neighbours.

However, because E occurs on 2022-06-20, which is a Monday, and the value on a Monday is expected to be high according to the learned weekly pattern depicted in Fig 2 (c), E is deemed insufficiently high by the model and hence classified as an anomaly.

Subsequently, we coined the phrase "in-season anomalies" to refer to anomalies that fall within their weekly range but are "off" in terms of their weekday position. The points of A, C, D, and E in Fig 3 are in-season anomalies. It is evident that many in-season anomalies do not look obvious and in reality, they may be false positives as we have discussed with stakeholders in real projects.

Consequently, the issue we face is how to suppress spurious in-season anomalies while preserving genuine ones. The problem is prevalent in business applications. Cellphone usage and revenue streams, for instance, exhibit a pronounced weekly pattern, and in-season anomalies are widespread.

Nonetheless, there remains a shortage of research in this field. We have not yet discovered any pertinent discussions on the subject.

### III. METHODOLOGY

#### A. k-parameter approach

To solve the problem, i.e. how to suppress the in-season anomalies, we propose an approach by using a k-parameter to provide additional tolerance for in-season anomalies. The fundamental concept is to modify the predicted value, thus reducing the residual, hence suppressing the anomaly detection to an ideal extent. The approach is implemented using the following formulas.

$$v_p = trend \cdot season \tag{2}$$

$$v_{max} = trend \cdot \max(season) \tag{3}$$
$$v_{min} = trend \cdot \min(season)$$

$$r_{0,high} = \frac{v_p}{v_{max}} \tag{4}$$
$$r_{0,low} = \frac{v_{min}}{v_p} \tag{5}$$

$$r_{k,high} = r_{0,high}^{1-k} \tag{6}$$
$$r_{k,low} = r_{0,low}^{1-k} \tag{7}$$

$$v'_{p,high} = v_{max} \cdot r_{k,high} \tag{8}$$
$$v'_{p,low} = v_{min} \cdot r_{k,low}$$

$$resid'_{high} = \frac{v}{v'_{p,high}}$$
$$resid'_{low} = \frac{v}{v'_{p,low}}$$

$$thresh = n \cdot \text{std}(resid)$$

$$p_{high} = \frac{resid'_{high}}{1+resid} \tag{9}$$
$$p_{low} = \frac{\frac{1}{1+thresh}}{resid'_{low}}$$

$$p = \max(p_{high}, p_{low})$$

$$class = 0, if\ p \leq 1$$
$$class = 1, if\ otherwise$$

where

$v_{max}$ is the predicted weekly maximum value;
$v_{min}$ is the predicted weekly minimum value;

$r_{0,high}$ is the ratio between the predicted value and the predicted weekly maximum value;
$r_{0,low}$ is the ratio between the predicted weekly minimum value and the predicted value;

$r_{k,high}$ is the adjusted ratio of $r_{0,high}$;

$r_{k,low}$ is the adjusted ratio of $r_{0,low}$;

$k$ is the arbitrary parameter introduced by the approach. It can be set in the range of [0, 1];

$v'_{p,high}$ is the adjusted predicted value that is used to calculate $resid'_{high}$;
$v'_{p,low}$ is the adjusted predicted value that is used to calculate $resid'_{low}$;

$resid'_{high}$ is the adjusted residual that is used to evaluate the extent to which the actual value is too high;
$resid'_{low}$ is the adjusted residual that is used to evaluate the extent to which the actual value is too low;

$p_{high}$ is the standardized estimate of the extent to which the actual value is too high;
$p_{low}$ is the standardized estimate of the extent to which the actual value is too low;

$p$ is the final standardized estimate of the extent to which the actual value is an anomaly. $p$ is in the range of [0, inf).

The fundamental technique of the approach is represented in (4)-(7). It introduces a parameter $k$, which can be arbitrarily set in the range of [0, 1]. Due to $k$, the approach converts the predicted value $v_p$ from a single value to two values: $v'_{p,high}$ and $v'_{p,low}$. $v'_{p,high}$ serves as a baseline for measuring the extent to which the actual value is too high, and $v'_{p,low}$ for too low.

Since $k$ is in the range of [0, 1], and $r_{0,high}$ is in the range of (0, 1], it follows from (6) that

$$r_{k,high} \geq r_{0,high}$$

Also, from (4), (6) and (8) we obtain

$$v_p = v_{max} \cdot r_{0,high}$$
$$v'_{p,high} = v_{max} \cdot r_{k,high} \geq v_{max} \cdot r_{0,high} = v_p \tag{10}$$
$$r_{0,high} \leq 1$$
$$r_{k,high} \leq 1$$
$$v'_{p,high} \leq v_{max}$$

Together with (10) we obtain

$$v_p \leq v'_{p,high} \leq v_{max}$$
$$resid'_{high} = \frac{v}{v'_{p,high}} \leq \frac{v}{v_p} = resid$$

Therefore, due to the reduced residual $resid'_{high}$, the model will obtain a reduced $p_{high}$ for in-season anomalies from (9). Similarly, on the side of the low threshold, the reduced residual $resid'_{low}$ will cause a decreased $p_{low}$. Consequently, the model will suppress the in-season anomalies to a degree controlled by the k-parameter.

In fact, Fig 4 depicts the controlling effect of the k-parameter. The greater the value of $k$, the greater the values of $r_{k,high}$ and $r_{k,low}$; as a result, more adjustment will be applied to $v_p$, thus more reduction to $p$, therefore suppressing the in-season anomaly to a greater degree, and vice versa.

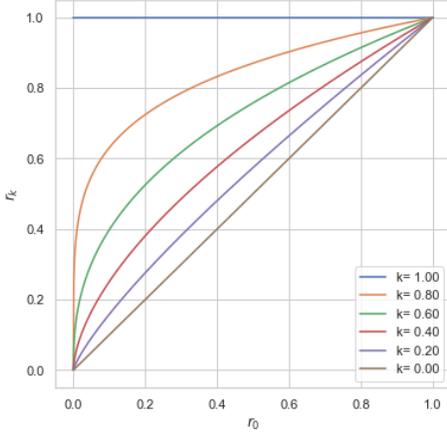

Fig 4. k-parameter effect. k is arbitrarily set in the range of [0, 1]. The greater the value of k, the greater the $r_k$, hence more adjustment will be applied to $v_p$, thus more reduction to p, therefore suppressing the in-season anomaly to a greater degree, and vice versa.

In the meanwhile, it is evident from (8) that the adjustment is in a limited extent. The greatest impact (when $k = 1$) is causing

$$v'_{p,high} = v_{max}$$
$$v'_{p,low} = v_{min}$$

In this circumstance, all of the modified predicted values will be identical to the season's maximum or minimum. In other words, seasonality will be entirely ignored, and the decision boundary in Fig 3 will become parallel to the trend, i.e. trend plus or minus a constant value. When $k$ is set between 0 and 1, the decision boundary will be more fluctuating.

### B. Overall effect of z and k parameters

Consequently, Fig 5 summarizes the overall effect of both $z$ and $k$ parameters. $k$ is used to manage in-season anomalies, whereas $z$ is used to manage overall (i.e. non-in-season) anomalies.

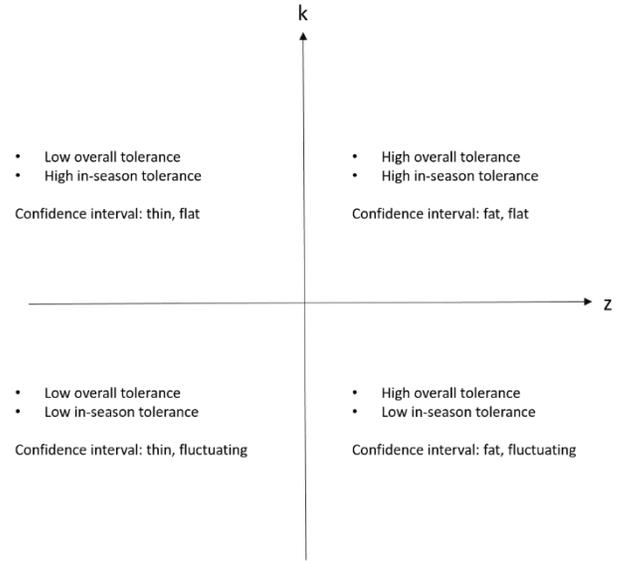

Fig 5. Combinational effect of z and k parameters. k is used to manage in-season anomalies, whereas z is used to manage overall (i.e. non-in-season) anomalies.

## IV. RESULT

### A. Enhanced result

Using the k-parameter approach, Fig 6 demonstrates an improvement in anomaly detection. $k$ is set to 0.40. Numerous detected in-season anomalies in Fig 3, including A, D, and E, have been suppressed. On the other hand, extremely strong in-season anomalies, such as C, and non-in-season anomalies, such as B, are retained. Thus, genuine anomalies are preserved.

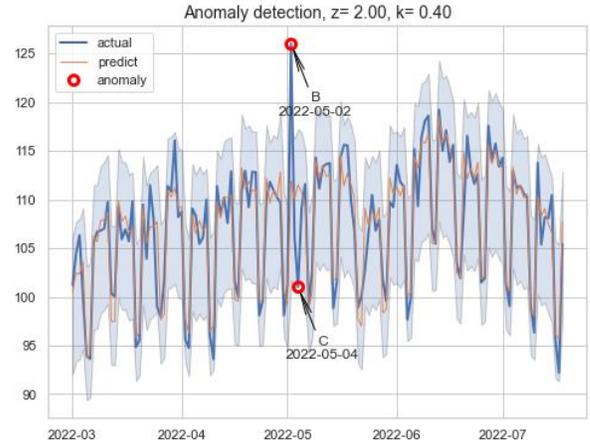

Fig 6. An improvement in anomaly detection using the k-parameter approach. k is set to 0.40. Numerous detected in-season anomalies in Fig 3, including A, D, and E, have been suppressed. On the other hand, extremely strong in-season anomalies, such as C, and non-in-season anomalies, such as B, are retained. Thus, genuine anomalies are preserved.

### B. Parameter selection

Consequently, using the two parameters $z$ and $k$, we can handle the anomaly detection task with flexibility. The effects of various combinations of $z$ and $k$ are depicted in Fig 7. In fact, Fig 7 is an example of Fig 5 in reality.

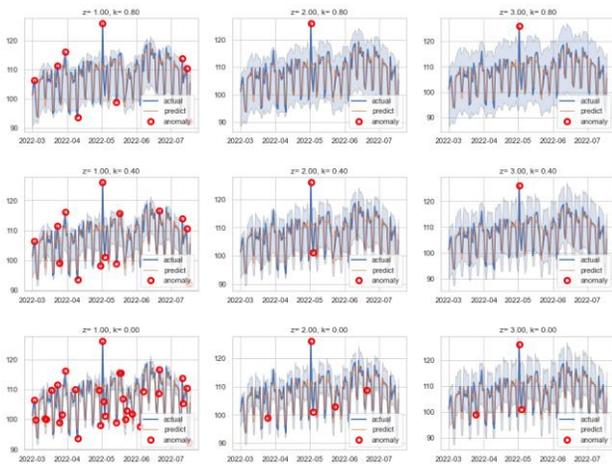

Fig 7. Effects of various combinations of z and k. It is an example of Fig 5 in reality.

Lastly, how to select the parameters $z$ and $k$? Our answer is to engage subject matter experts. That is, we learn the true anomalies confirmed by subject matter experts, then select the closest chart from Fig 7, and there we can determine the parameters accordingly. This way effectively converts the unsupervised learning process into a supervised learning process, taking advantage of domain knowledge for supervision.

## V. CONCLUSION

False in-season anomaly is an issue in daily time series anomaly detection using decomposition method. We propose a k-parameter approach to solve the problem by providing more tolerance for in-season anomalies. The adjustment intensity is controllable. The approach obtain good results suppressing false in-season anomalies whereas retaining true anomalies.

Further improvements may include the study of applying the k-parameter approach to anomaly detection techniques other than the decomposition method. For example, apply the approach to ARIMA method (capable of clustering seasonality patterns [10]), LDA (Latent Dirichlet Allocation, a variant of seasonal-trend decomposition method [11]), and multi-series anomaly detection.